\documentclass[11pt]{article}

\usepackage[preprint]{acl}

\usepackage{times}
\usepackage{latexsym}

\usepackage[T1]{fontenc}

\usepackage[utf8]{inputenc}

\usepackage{microtype}

\usepackage{inconsolata}

\usepackage{graphicx}

\usepackage{amsmath}
\usepackage[whole]{bxcjkjatype}
\usepackage{multirow}
\usepackage{subcaption}
\usepackage{booktabs}
\usepackage{pifont}
\usepackage{xcolor}

\newcommand{\correct}{\textcolor{green!60!black}{\ding{51}}}
\newcommand{\incorrect}{\textcolor{red!70!black}{\ding{55}}}

\newcommand{\subfigwidth}{0.329\textwidth}

\usepackage[sort]{cleveref}
\crefname{equation}{Equation}{Equations}
\crefname{figure}{Figure}{Figures}
\crefname{table}{Table}{Tables}
\crefname{section}{\S}{\S}
\crefname{subtable}{Table}{Tables}

\crefname{subfigure}{Figure}{Figures}

\title{Revisiting Non-Verbatim Memorization in Large Language Models: \\The Role of Entity Surface Forms}

\author{
Yuto Nishida$^{1,2}$ $\;\;\;$ 
Naoki Shikoda$^{1}$ $\;\;\;$ 
Yosuke Kishinami$^{2}$ $\;\;\;$ 
Ryo Fujii$^{2}$ $\;\;\;$ 
\\ 
\textbf{
Makoto Morishita$^{2}$ $\;\;\;$ 
Hidetaka Kamigaito$^{1}$ $\;\;\;$ 
Taro Watanabe$^{1}$ $\;\;\;$ 
} \\
$^1$Nara Institute of Science and Technology $\;$ 
$^2$Future Corporation $\;$ \\
\texttt{\{nishida.yuto.nu8, kamigaito.h, taro\}@is.naist.jp} \\
\texttt{shikoda.naoki.sm1@naist.ac.jp} \\
\texttt{\{y.kishinami.rh, r.fujii.6d, m.morishita.pi\}@future.co.jp}
}

\begin{document}
\maketitle

\begin{abstract}
Understanding what kinds of factual knowledge large language models (LLMs) memorize is essential for evaluating their reliability and limitations.
Entity-based QA is a common framework for analyzing non-verbatim memorization, but typical evaluations query each entity using a single canonical surface form, making it difficult to disentangle fact memorization from access through a particular name.
We introduce \emph{RedirectQA},\footnote{\url{https://huggingface.co/datasets/naist-nlp/RedirectQA}} an entity-based QA dataset that uses Wikipedia redirect information to associate Wikidata factual triples with categorized surface forms for each entity, including alternative names, abbreviations, spelling variants, and common erroneous forms.
Across 13 LLMs, we examine surface-conditioned factual memorization and find that prediction outcomes often change when only the entity surface form changes.
This inconsistency is category-dependent: models are more robust to minor orthographic variations than to larger lexical variations such as aliases and abbreviations.
Frequency analyses further suggest that both entity- and surface-level frequencies are associated with accuracy, and that entity frequency often contributes beyond surface frequency.
Overall, factual memorization appears neither purely surface-specific nor fully surface-invariant, highlighting the importance of surface-form diversity in evaluating non-verbatim memorization.
\end{abstract}

\begin{table*}[t]
    \small
    \centering
    \tabcolsep 3pt
    \begin{tabular}{@{}llcllc@{}}
        \toprule
        & \multicolumn{1}{c}{Surface Category} & \multicolumn{1}{c}{Type} & \multicolumn{1}{c}{Question} & LLM Answer & Eval. \\
        \midrule

        \multirow{2}{*}{Case 1}
        & Canonical & - & \textit{What is \textbf{David Guetta}'s occupation?} & \textit{DJ} & \correct \\
        & \texttt{from pseudonyms} & Alt./Abbrev. & \textit{What is \textbf{Jack Back}'s occupation?} & \textit{actor} & \incorrect \\
        \midrule[.03em]

        \multirow{2}{*}{Case 2}
        & Canonical & - & \textit{What sport does \textbf{José García Castro} play?} & \textit{baseball} & \incorrect \\
        & \texttt{from alternative names} & Alt./Abbrev. & \textit{What sport does \textbf{Pepillo II} play?} & \textit{soccer} & \correct \\
        \midrule[.03em]

        \multirow{2}{*}{Case 3}
        & Canonical & - & \textit{In what city was \textbf{J. M. Coetzee} born?} & \textit{Cape Town} & \correct \\
        & \texttt{from modifications} & Spell. Var. & \textit{In what city was \textbf{J M Coetzee} born?} & \textit{Cape Town} & \correct \\
        \midrule[.03em]

        \multirow{2}{*}{Case 4}
        & Canonical & - & \textit{What is \textbf{Stan Coveleski}'s occupation?} & \textit{baseball player} & \correct \\
        & \texttt{from misspellings} & Typ. Err. & \textit{What is \textbf{Stan Covaleski}'s occupation?} & \textit{actor} & \incorrect \\
        \bottomrule
    \end{tabular}
\caption{
Illustrative examples of surface-conditioned factual access in RedirectQA.
Each pair of rows refers to the same Wikidata entity and factual triple; only the subject entity surface form in the question is changed.
The gold answer is therefore fixed within each case, but the Pythia-12B predictions can flip between \correct correct and \incorrect incorrect.
The examples show canonical-to-redirect failures, the reverse pattern, robustness to a minor orthographic variant, and fragility to a common misspelling.
Aggregate results across 13 LLMs are reported in \cref{sec:experiments}.
}
\label{tab:example}
\end{table*}

\section{Introduction}
\label{sec:introduction}

Large language models (LLMs) store a wide range of factual knowledge in their parameters, enabling them to answer many knowledge-intensive questions without external retrieval~\cite{petroni-etal-2019-language,yu2023generate}.
At the same time, when the required knowledge is absent or inaccessible, LLMs may produce hallucinated or erroneous answers~\cite{simhi2024distinguishingignoranceerrorllm}.
Understanding what factual knowledge LLMs memorize non-verbatim, and under what conditions they can access it, is therefore central to evaluating their reliability and limitations.

A common way to analyze non-verbatim memorization is entity-based question answering (QA), where models are queried about factual relations involving entities and memorization is measured by answer accuracy~\cite{sciavolino-etal-2021-simple,mallen-etal-2023-trust,maekawa-etal-2024-retrieval}.
This line of work has shown that facts about low-frequency or low-popularity entities are less likely to be memorized~\cite{kandpal2023large,mallen-etal-2023-trust,maekawa-etal-2024-retrieval}.
However, typical evaluations instantiate each entity using a single canonical surface form.
This makes it difficult to disentangle whether a model has memorized a fact about an entity from whether it can access that fact through the particular name used in the question.

This distinction matters because entities are often referred to by multiple surface forms.
A model that answers correctly for a canonical name such as \emph{Pelé} may not necessarily access the same fact when the entity is referred to as \emph{Edson Arantes do Nascimento}.
Indeed, in our preliminary diagnostic using Pythia-12B~\cite{biderman2023pythia} on a redirect-augmented version of PopQA~\cite{mallen-etal-2023-trust}, 23.7\% of canonical--redirect question pairs yield inconsistent predictions (Appendix~\ref{appendix:preliminary_experiment}).
This observation motivates a systematic evaluation in which the underlying fact is controlled while the entity surface form is varied.

To analyze this phenomenon systematically, we introduce \emph{RedirectQA}, an entity-based QA dataset that associates Wikidata factual triples with multiple entity surface forms using Wikipedia redirect information.
The key design of RedirectQA is to hold the factual relation and gold answer fixed while varying only the surface form of the subject entity.
As illustrated in \cref{tab:example}, this design exposes cases where a model answers correctly under one surface form but incorrectly under another, even though the underlying fact is unchanged.
Redirect surface forms are further annotated with categories such as alternative names, abbreviations, spelling variants, and common erroneous forms, enabling controlled analyses of how different types of naming variation affect factual QA.

Using RedirectQA, we evaluate 13 LLMs and find that prediction outcomes often differ across surface forms of the same entity, even though the underlying factual triple is held fixed.
The inconsistency is category-dependent: models are relatively robust to minor orthographic variations, such as spelling differences, diacritics, and punctuation changes, but are less consistent for larger lexical variations, such as aliases, alternative names, and abbreviations.
These results indicate that non-verbatim memorization cannot be treated as fully surface-invariant, even when the entity and fact remain the same.

We further analyze how entity- and surface-level frequencies relate to memorization.
By decomposing aggregate entity frequency into surface-level frequencies, we find that accuracy is associated with both the frequency of a specific surface form and the aggregate frequency of the corresponding entity, with entity frequency often contributing beyond surface frequency.
This pattern suggests cross-surface coupling in factual access, rather than purely independent memorization of each surface form.
Together with the consistency results, these findings point to an intermediate picture in which factual memorization is neither purely surface-specific nor fully surface-invariant.

Overall, our work shows that evaluating non-verbatim memorization through canonical entity names alone can miss surface-conditioned failures in factual access.
RedirectQA provides a controlled resource for studying these effects, highlighting surface-form diversity as a key factor in evaluating what LLMs memorize and how reliably they can access it.

\section{RedirectQA}
\label{sec:dataset}

We introduce \emph{RedirectQA}, an entity-based factual QA dataset designed to analyze how LLMs access the same factual knowledge through different surface forms of an entity.
RedirectQA associates Wikidata factual triples in the form of $(\textit{subject}, \textit{relation}, \textit{object})$ with multiple subject entity surface forms using Wikipedia redirect information.
The key design is to keep the factual relation and gold answer fixed while varying the surface form of the subject entity.
We follow the open-domain QA setting~\citep{roberts-etal-2020-much}, evaluating models on factual questions without providing external evidence.

\subsection{Wikipedia Redirects as Surface-Form Resources}
\label{sec:wikipedia_redirects}

Wikipedia article titles are chosen according to naming guidelines,\footnote{\url{https://en.wikipedia.org/wiki/Wikipedia:Article_titles}} typically favoring recognizable, natural, and searchable expressions among possible names for a topic or entity.
To make articles accessible through alternative expressions, Wikipedia provides redirect pages, which automatically forward users from a redirect title to the corresponding main article.
For example, the page titled ``NYT'' redirects to the article ``The New York Times.''
Such redirects provide a large-scale source of surface forms that refer to the same underlying entity.

Redirect pages are often annotated with redirect categories that describe the relationship between the redirect title and the main article title.\footnote{A redirect page may have zero or multiple categories.}
For instance, the redirect page ``NYT'' is annotated with \texttt{Redirects from initialisms}, indicating that ``NYT'' is an initialism for ``The New York Times.''\footnote{Hereafter, we omit the prefix \texttt{Redirects} when referring to category names.}
These categories allow us to group surface forms by the type of variation they represent.

In RedirectQA, we use this redirect structure to define two types of subject entity surface forms.
The \emph{canonical surface form} is the article title associated with the entity, while \emph{redirect surface forms} are the titles of pages that redirect to that article.

However, not all redirects correspond to genuine surface-form variants of the target entity.
For example, in the category \texttt{from books}, the title of a book may redirect to the article of its author, rather than to an alternative name for the same entity.
We therefore manually selected 33 frequent redirect categories that clearly represent surface-form variation.
We group the selected categories into three broad types.
First, \emph{Alternative Names and Abbreviations} include cases such as ``Stevland Hardaway Judkins'' redirecting to ``Stevie Wonder'' (\texttt{from birth names}).
Second, \emph{Spelling Variants} include cases such as ``Nicolas Sarközy'' redirecting to ``Nicolas Sarkozy'' (\texttt{from titles with diacritics}).
Third, \emph{Typical Errors} include cases such as ``Christian Ronaldo'' redirecting to ``Cristiano Ronaldo'' (\texttt{from incorrect names}).
The selected categories and their types are listed in \cref{tab:redirect_category_with_count}.

\subsection{Dataset Structure}
\label{sec:dataset_structure}

For each factual triple, RedirectQA creates instances in which the subject entity is expressed using different surface forms while the relation and gold answer remain fixed.

We use three dataset units throughout the paper.
A \emph{surface-form instance}, or simply a \emph{surface instance}, pairs a factual triple with a subject surface form.
A \emph{canonical--redirect pair} consists of a redirect surface instance and the corresponding canonical surface instance for the same factual triple.
This pair is the unit used in our consistency analyses.
A \emph{question realization} is obtained by rendering a surface instance with a relation-specific question template.

\subsection{Dataset Construction}
\label{sec:dataset_construction}

\begin{figure}[t]
\centering
\includegraphics[width=1.0\linewidth]{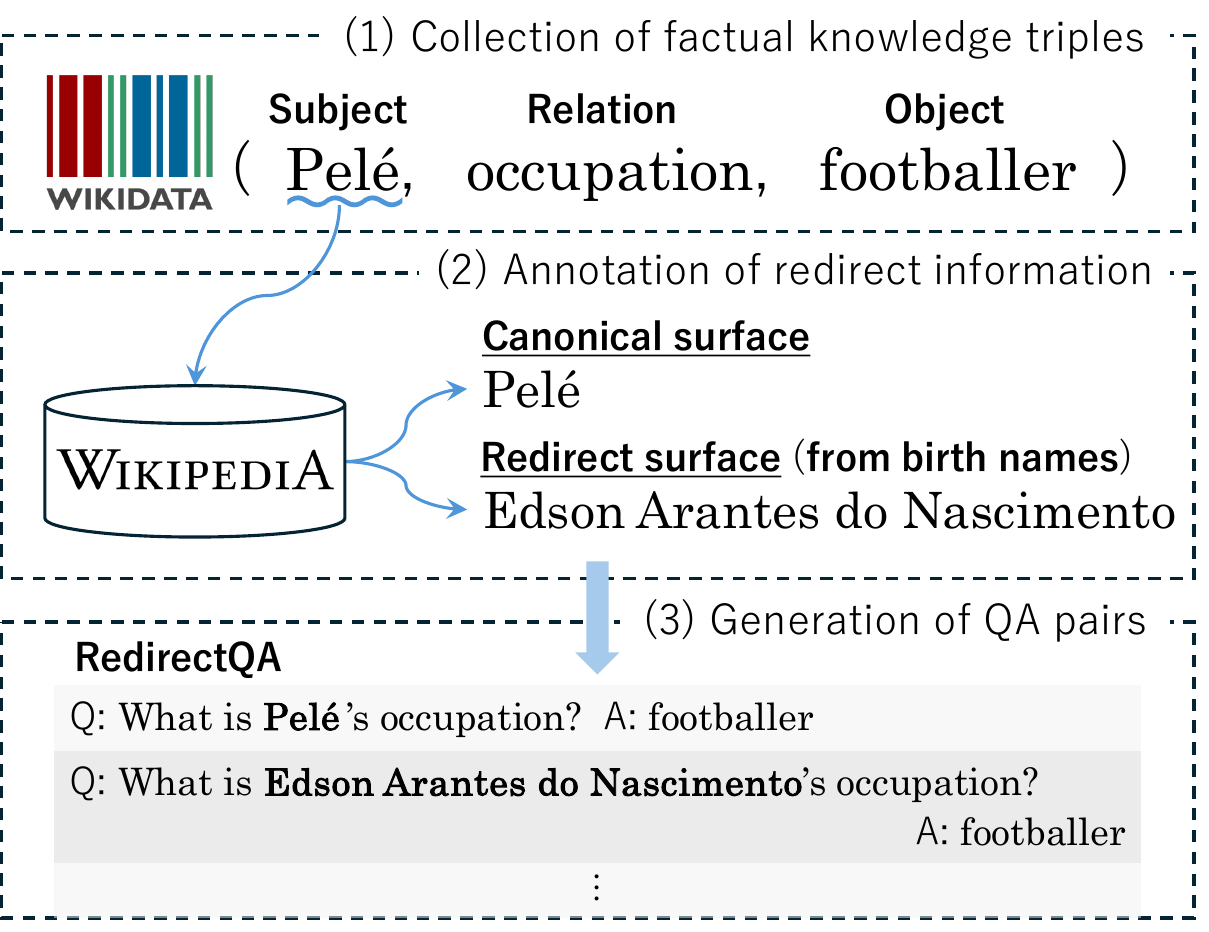}
\caption{
Overview of the RedirectQA construction process:
(1) Factual triples are collected from Wikidata.
(2) Each subject entity is associated with canonical and redirect surface forms, together with redirect categories, using Wikipedia redirects.
(3) Question realizations are generated from surface instances using relation-specific question templates.
}
\label{fig:redirectqa}
\end{figure}

The overall construction process is illustrated in \cref{fig:redirectqa}.
We first collect factual triples from Wikidata, then associate each subject entity with canonical and redirect surface forms using Wikipedia redirect information, and finally render surface instances into question realizations with relation-specific templates.

\paragraph{(1) Collection of factual triples.}
We collected factual triples from a Wikidata dump, targeting entities with English labels and restricting the relation types to 16 (e.g., \texttt{occupation}), following the setup of \citet{mallen-etal-2023-trust}.
To ensure that each factual question has a unique and unambiguous gold answer, we excluded cases where multiple English entities shared the same canonical surface form in Wikidata.
We also filtered out triples without corresponding Wikipedia pages, as well as triples whose subject or object entities had zero pageviews over the past year.\footnote{We used Wikimedia pageview statistics aggregated over 2024-01--2024-12.}
Finally, we randomly sampled 500k triples from the remaining set for subsequent processing.

\paragraph{(2) Annotation of redirect information.}
For each subject entity in the sampled triples, we collected redirect surface forms and their redirect categories from Wikipedia.
We discarded redirect surface forms whose categories were not among the selected categories described in \cref{sec:wikipedia_redirects}, and removed triples for which no valid redirect surface remained.
To reduce ambiguity and duplication, we further removed redirect surfaces whose strings matched existing English entity labels in Wikidata.
Finally, to mitigate severe class imbalance, we downsampled surface instances from overrepresented categories such as \texttt{from titles without diacritics} and \texttt{from other capitalisations}.
This balancing step reduces the dominance of a small number of redirect types while maintaining approximately 30k surface instances.

\paragraph{(3) Generation of question realizations.}
For each surface instance, we generated questions using relation-specific templates.
To reduce sensitivity to question wording, we used two templates for each relation type, following prior evidence that LLM predictions can be sensitive to superficial variations in question templates~\cite{sakai-etal-2024-toward}.
The first is the original template used by \citet{mallen-etal-2023-trust}.
The second is a paraphrase of the original template generated using GPT-4o~\cite{openai2024gpt4o}, designed to preserve the same factual semantics while differing in question wording.
Thus, each surface instance is rendered into two question realizations.

\paragraph{Dataset Statistics.}
After these steps, RedirectQA contains 30,560 surface instances derived from 14,672 factual triples: 14,672 canonical surface instances and 15,888 redirect surface instances.
The 15,888 redirect surface instances define the canonical--redirect pairs used in our consistency analyses.
Because each surface instance is rendered with two templates, the dataset contains 61,120 question realizations in total.
Among the redirect surface instances, 8,667 are associated with \emph{Alternative Names and Abbreviations}, 4,928 with \emph{Spelling Variants}, and 2,884 with \emph{Typical Errors}.\footnote{
These types are not mutually exclusive, as a redirect surface instance may be associated with multiple categories; therefore, the type-level counts do not sum to the total number of instances.
}
A detailed breakdown of redirect categories and their surface-instance counts is shown in \cref{tab:redirect_category_with_count}.

\begin{figure*}[t]
\centering
\includegraphics[width=1.0\linewidth]{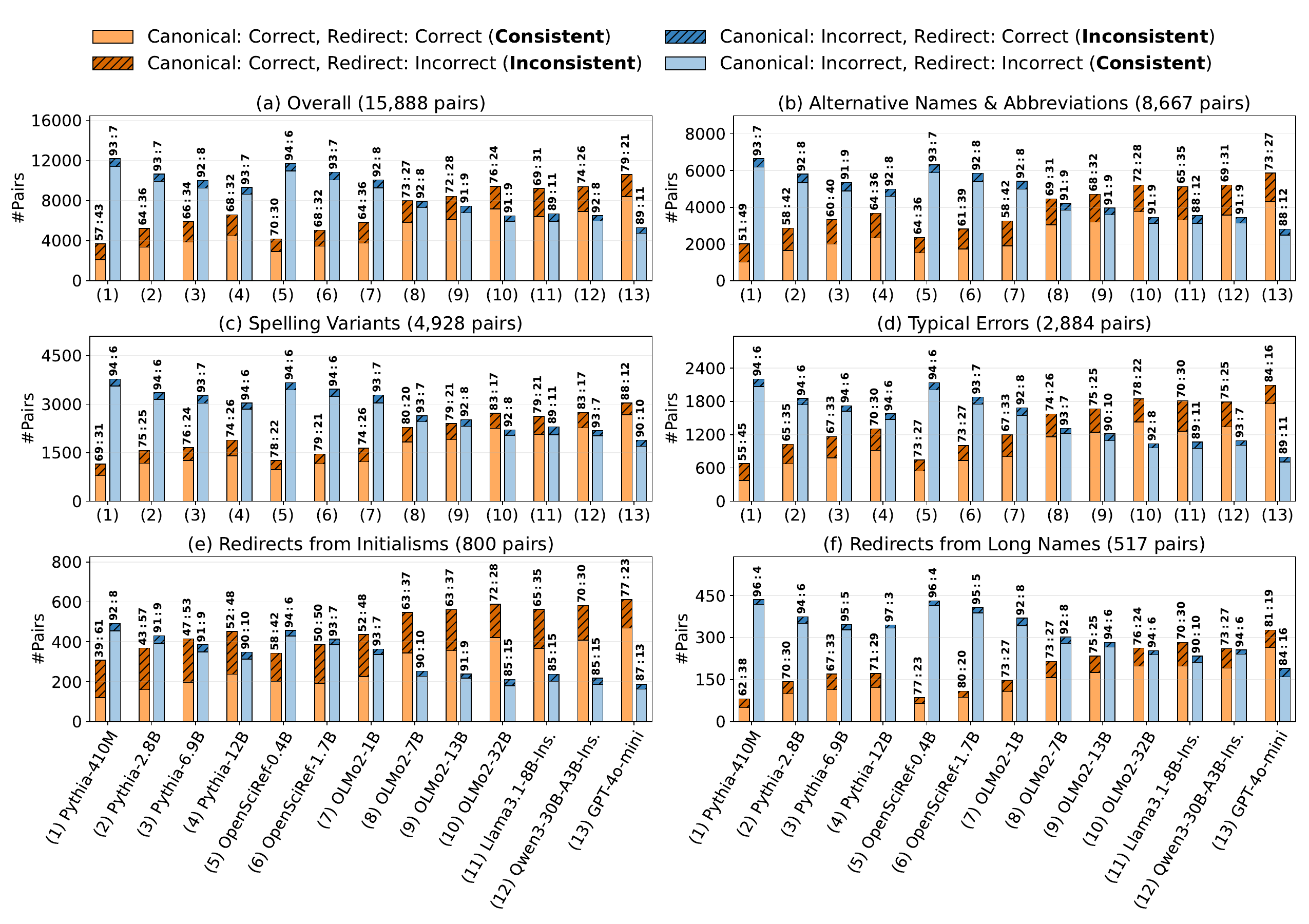}
\caption{
Prediction consistency between canonical and redirect surface forms on RedirectQA using the original question template.
Each panel reports results for a redirect type or selected redirect category.
For each model, the left stacked bar contains canonical--redirect pairs where the canonical question is answered correctly, and the right stacked bar contains pairs where it is answered incorrectly.
Light segments indicate consistent correctness outcomes across the two surface forms, while dark hatched segments indicate correctness flips.
Numbers above bars show the consistent:inconsistent percentage split within each bar.
}
\label{fig:redirectqa_summary}
\end{figure*}

\section{Experiments}
\label{sec:experiments}
This section evaluates whether factual QA behavior remains consistent when only the subject entity surface form is changed.
We first describe the evaluated models and inference protocol, and then analyze prediction consistency across canonical--redirect pairs and redirect categories.
Frequency-based analyses are presented in \cref{sec:analysis}.

\subsection{Experimental Setup}
\label{sec:experimental_setup}

\paragraph{Models.}
We evaluated 13 LLMs spanning three tiers of accessibility and training transparency:
\emph{transparent models} with well-documented pretraining data and procedures, \emph{open-weight models} with publicly available weights but limited training transparency, and a \emph{proprietary model} accessed via an API.
This design supports corpus-based frequency analyses that require traceable pretraining corpora, such as the analysis in \cref{sec:analysis}, while also testing whether surface-form effects persist across a broader range of widely used models.

We used three families of transparent models.
For Pythia~\cite{biderman2023pythia}, we used four model sizes: 410M, 2.8B, 6.9B, and 12B, pretrained on the Pile~\cite{gao2020pile800gbdatasetdiverse}.
For OpenSciRef v0.01~\cite{nezhurina2025opensciref001openreproduciblereference}, we used the Pile-pretrained variants at 0.4B and 1.7B among its publicly released corpus-specific variants.
For OLMo~2~\cite{olmo20242olmo2furious}, we used the final Stage-1 checkpoints at 1B, 7B, 13B, and 32B.
OLMo~2 base-model training consists of Stage~1 pretraining on OLMo Mix 1124 followed by Stage~2 mid-training.
Because our frequency analyses target the pretraining corpus, we evaluate the final Stage-1 checkpoints rather than checkpoints after Stage~2.
These Stage-1 checkpoints share the same data mixture, although their training budgets differ across model sizes.

To include strong instruction-tuned open-weight models, we evaluated Qwen~3~\cite{yang2025qwen3technicalreport} 30B-A3B-Instruct and Llama~3.1~\cite{grattafiori2024llama3herdmodels} 8B-Instruct.
As a representative \emph{proprietary} model, we used the GPT-4o-mini~\cite{openai2024gpt4omini} snapshot \texttt{gpt-4o-mini-2024-07-18} via the API.

\paragraph{Inference and Evaluation.}
For local inference on training-transparent and open-weight models, we applied 8-bit quantization to reduce memory usage.
Following \citet{mallen-etal-2023-trust}, we used prompts of the form ``\texttt{Q: <question> A:}'' in a 15-shot setting.
For each test question, the demonstrations were deterministically sampled with a fixed random seed from canonical-surface instances of other relation types, excluding the same factual triple.
Specifically, we sampled one demonstration from each of the other 15 relation types.

For local models, we generated up to 15 new tokens and extracted the first generated line as the prediction.
For GPT-4o-mini, we used the API with temperature 0, top-p 1, and a maximum of 100 output tokens, applying the same first-line extraction.
We evaluated predictions using alias-aware string matching.
For each question, a prediction was counted as correct if the extracted prediction contained any acceptable surface form of the gold answer entity, allowing simple case variants.
This avoids penalizing alternative valid names of the answer entity when they are included in the acceptable surface set, while retaining a string-based evaluation appropriate for our entity-answering setting.

\subsection{Prediction Consistency Across Surface-Form Categories}
\label{sec:consistency}
We analyze whether model predictions remain consistent when only the subject entity surface form is changed.
For each canonical--redirect pair, we compare the correctness of the model's answer under the canonical surface form with that under the corresponding redirect surface form.
We call a pair \emph{consistent} if the two predictions have the same correctness outcome, i.e., both are correct or both are incorrect, and \emph{inconsistent} otherwise.
Because correct--correct and incorrect--incorrect consistency have different interpretations, we separately analyze pairs where the canonical question is answered correctly and pairs where it is answered incorrectly.

\Cref{fig:redirectqa_summary} summarizes prediction consistency across 13 LLMs using the original question template.
Overall, surface-form changes induce non-negligible correctness flips across all model classes.
Within several model families, larger models tend to be more consistent, but the effect is not monotonic across all models or categories.
Moreover, even strong instruction-tuned and proprietary models do not achieve perfect consistency, indicating that access to factual knowledge remains sensitive to how the subject entity is named.

The category-wise results reveal systematic differences.
\emph{Spelling Variants} yield the highest consistency across models, suggesting that models are relatively robust to minor orthographic changes such as punctuation, capitalization, and diacritics.
By contrast, \emph{Alternative Names and Abbreviations} show substantially lower consistency, indicating that larger lexical changes are more likely to disrupt factual access.
\emph{Typical Errors} generally fall between these two types, reflecting partial but imperfect robustness to misspellings, miscapitalizations, and incorrect names.

The selected subcategories within \emph{Alternative Names and Abbreviations} further illustrate that not all lexical variants are equally difficult.
Redirects from initialisms are especially challenging: abbreviated forms such as \emph{NYT} for \emph{The New York Times} often fail to elicit the same answer as the canonical surface form.
In contrast, redirects from long names tend to be more consistent, possibly because some longer alternative names preserve lexical or semantic cues that support factual access.
These trends show that surface-form effects are not merely random noise, but depend on the type of relation between the redirect and canonical surface forms.

We repeat the same analysis using the paraphrased question template generated by GPT-4o and report the results in Appendix~\ref{appendix:template_robustness}.
Although absolute accuracy can vary with question wording, the model-wise consistency patterns and category-wise differences largely mirror those obtained with the original template.
This supports the conclusion that the observed surface-form effects are not artifacts of a single question template.

The illustrative examples in \cref{tab:example} provide concrete instances of these aggregate patterns, including canonical-to-redirect failures, the reverse pattern, robustness to a minor orthographic variant, and fragility to a common misspelling.
The reverse pattern is particularly informative: a model can fail under the Wikipedia canonical title but succeed under an alternative surface form, suggesting that human-oriented canonicality does not necessarily coincide with the surface form through which an LLM most reliably accesses a fact.
Thus, RedirectQA captures not only degradation from canonical to redirect surfaces, but also asymmetric surface dependence in factual access.

\section{Analysis: Entity- and Surface-Level Frequency Signals}
\label{sec:analysis}

\begin{figure*}[t]
\captionsetup[subfigure]{justification=centering, skip=3pt}
  \centering
  
  \begin{minipage}[b]{\subfigwidth}
    \centering
    \includegraphics[width=\textwidth]{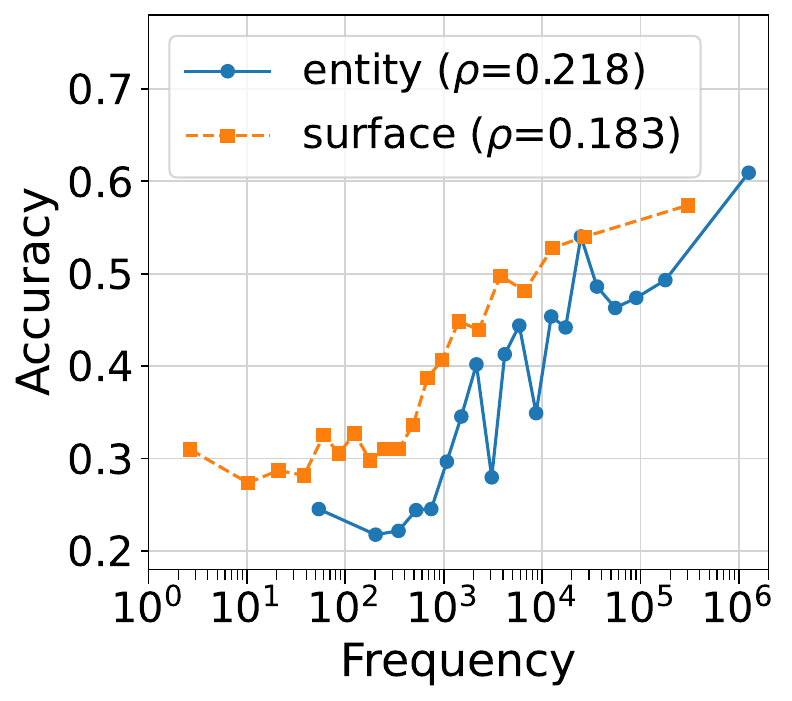}
    \subcaption{Overall (4,284 instances)}
    \label{fig:count_vs_score_overall}
  \end{minipage}
  \hfill
  \begin{minipage}[b]{\subfigwidth}
    \centering
    \includegraphics[width=\textwidth]{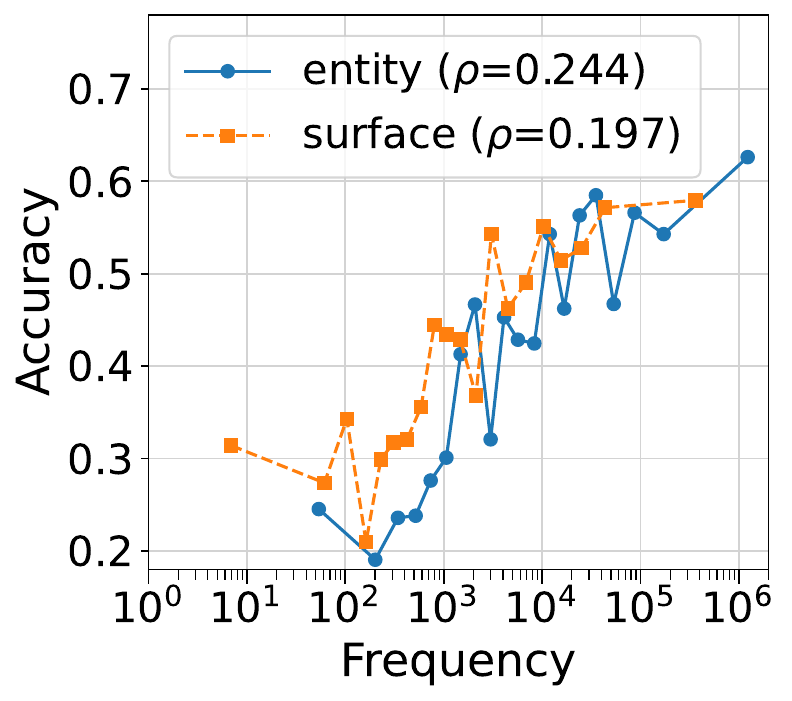}
    \subcaption{Canonical only (2,112 instances)}\label{fig:count_vs_score_canonical}
  \end{minipage}
  \hfill
  \begin{minipage}[b]{\subfigwidth}
    \centering
    \includegraphics[width=\textwidth]{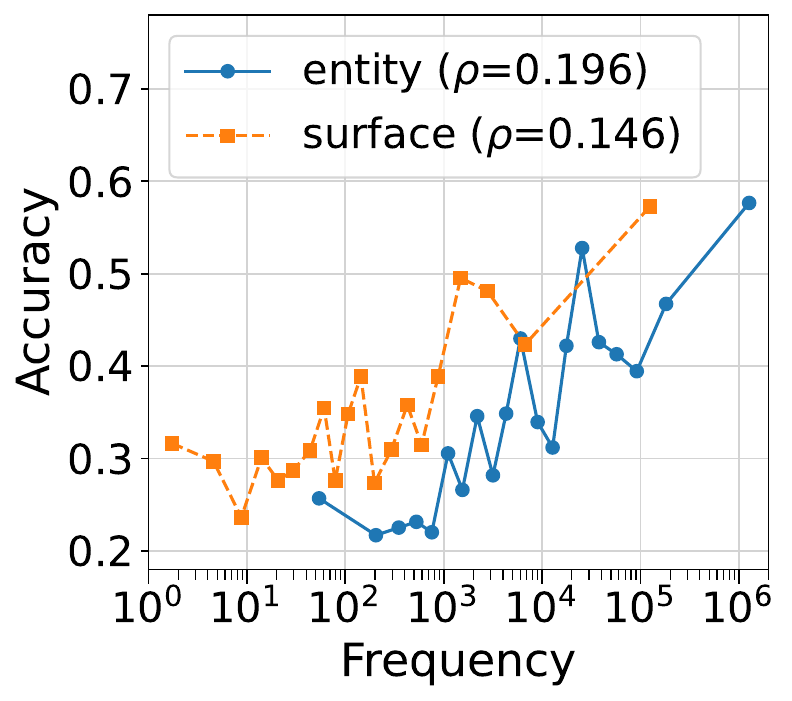}
    \subcaption{Redirect only (2,172 instances)}\label{fig:count_vs_score_redirect}
  \end{minipage}

\caption{
Relationship between accuracy and entity/surface frequencies for Pythia-12B.
Each point shows the mean accuracy of surface instances within one of 20 frequency bins with approximately equal numbers of instances.
For each surface instance, accuracy is averaged over the two question realizations.
Pearson correlations $\rho$ are computed between $\log(\text{frequency})$ and accuracy and are shown in the legend.
}
\label{fig:count_vs_score}
\end{figure*}

In \cref{sec:consistency}, we observed that factual QA predictions are not fully consistent across canonical and redirect surface forms, and that the degree of consistency varies across redirect categories.
These findings raise a question about the granularity of factual memorization: are surface forms memorized independently, or is factual access coupled across different surface forms of the same entity?
If accuracy for a target surface form is associated only with that surface form's own frequency, this would support a strongly surface-specific view.
If aggregate entity frequency also predicts accuracy beyond the target surface frequency, however, this would suggest cross-surface coupling in factual access.
We investigate this question by analyzing how entity-level and surface-level frequencies relate to factual QA accuracy.

Previous studies have shown that entity frequency is positively correlated with factual memorization, as reflected in factual QA accuracy~\cite{kandpal2023large,maekawa-etal-2024-retrieval}.
Such studies typically estimate entity frequency from pretraining or related corpora by using an entity linker to identify mentions of an entity across surface forms, and then treating the total number of linked mentions as the entity's frequency.
We decompose this aggregate entity frequency into surface-level frequencies, allowing us to ask whether accuracy is associated with the frequency of the target surface form itself or with the aggregate frequency of the corresponding entity.

\subsection{Counting Entity and Surface Frequencies}
Following \citet{kandpal2023large}, we counted entity and surface frequencies from the pretraining corpora of the training-transparent model families.
For Pythia and OpenSciRef v0.01, we used the Pile dataset~\cite{gao2020pile800gbdatasetdiverse}, which contains approximately 300B tokens.
For OLMo~2, we estimated frequencies from OLMo Mix 1124~\cite{olmo20242olmo2furious}, the Stage-1 data mixture, by randomly sampling 10\% of documents; this yields a corpus size roughly comparable to the Pile in total tokens.

We performed large-scale entity linking using DBpedia Spotlight~\cite{mendes2011dbpedia}, which links text spans to Wikipedia entities.\footnote{We retrieved the corresponding Wikipedia entities by resolving DBpedia URIs obtained from the linker through the official DBpedia SPARQL endpoint.}
For each entity, \emph{entity frequency} is the total number of linked mentions of that entity.
For a particular surface form, \emph{surface frequency} is the number of linked mentions of the same entity whose span exactly matches that surface form.
Thus, entity frequency aggregates over all observed linked surface forms of the entity, whereas surface frequency refers to the specific surface form used in a RedirectQA surface instance.

We annotated each RedirectQA surface instance with the entity frequency of its subject entity and the surface frequency of its subject surface form.
Following \citet{kandpal2023large}, we filtered out zero-frequency cases, which may reflect entity-linking failures or missing corpus coverage and cannot be used in log-frequency analyses.
We retained surface instances only when the subject entity was linked at least once and the target subject surface form was observed at least once as a linked mention of that entity.
Under this filtering criterion, the Pile-based analysis for Pythia and OpenSciRef v0.01 retains 4,284 surface instances from 2,112 factual triples, while the OLMo Mix 1124 analysis for OLMo~2 retains 4,356 surface instances from 2,147 factual triples.
These filtered subsets are used in the frequency analyses below.

\subsection{Correlation Analysis}
We first examine the relationship between frequency and factual QA accuracy in three subsets: \emph{overall}, \emph{canonical-only}, and \emph{redirect-only}.
The canonical-only and redirect-only subsets contain surface instances whose subject entity is expressed with the canonical and redirect surface forms, respectively, while the overall subset contains both.
For each surface instance, we compute accuracy as the mean correctness score across the two question realizations and use this continuous score in the correlation analyses.

\Cref{fig:count_vs_score} reports the results for Pythia-12B.
Each plot bins surface instances by frequency and shows the mean accuracy in each bin; Pearson correlations between $\log(\text{frequency})$ and accuracy are shown in the legends.
For Pythia-12B, both entity and surface frequencies are positively correlated with accuracy in all three subsets, with all reported correlations significantly different from zero ($p<0.01$).
This extends prior findings that entity frequency is predictive of factual QA accuracy~\cite{kandpal2023large} by showing that surface frequency is also positively associated with accuracy.
Entity frequency correlates more strongly with accuracy than surface frequency for Pythia-12B, and Appendix~\ref{appendix:frequency_correlation} confirms that this difference is significant for the canonical-only subset.
The appendix further shows that, across all training-transparent models and subsets, both frequency types have statistically significant positive correlations with accuracy, with entity frequency consistently stronger in the canonical-only subset.

\begin{table}[t]
\centering
\small
\tabcolsep 3pt
\begin{tabular}{@{}llrrr@{}}
\toprule
& & \multicolumn{3}{c}{Partial Correlation}\\
\cmidrule(l){3-5}
Size & Correlation Type & Overall & Canonical & Redirect \\
\midrule
\multicolumn{5}{c}{OLMo~2} \\
 \midrule
\multirow{2}{*}{1B} & $\rho(\text{Ent}, \text{Acc} \mid \text{Surf})$ & $0.132^{*}$ & $0.182^{*}$ & $0.125^{*}$ \\
 & $\rho(\text{Surf}, \text{Acc} \mid \text{Ent})$ & $0.049^{*}$ & $-0.069^{*}$ & $0.084^{*}$ \\
\midrule
\multirow{2}{*}{7B} & $\rho(\text{Ent}, \text{Acc} \mid \text{Surf})$ & $0.104^{*}$ & $0.127^{*}$ & $0.120^{*}$ \\
 & $\rho(\text{Surf}, \text{Acc} \mid \text{Ent})$ & $0.091^{*}$ & $-0.010$\phantom{$^{*}$} & $0.106^{*}$ \\
\midrule
\multirow{2}{*}{13B} & $\rho(\text{Ent}, \text{Acc} \mid \text{Surf})$ & $0.088^{*}$ & $0.102^{*}$ & $0.106^{*}$ \\
 & $\rho(\text{Surf}, \text{Acc} \mid \text{Ent})$ & $0.098^{*}$ & $0.010$\phantom{$^{*}$} & $0.111^{*}$ \\
\midrule
\multirow{2}{*}{32B} & $\rho(\text{Ent}, \text{Acc} \mid \text{Surf})$ & $0.064^{*}$ & $0.113^{*}$ & $0.065^{*}$ \\
 & $\rho(\text{Surf}, \text{Acc} \mid \text{Ent})$  & $0.083^{*}$ & $-0.032$\phantom{$^{*}$} & $0.114^{*}$ \\

\midrule
\multicolumn{5}{c}{OpenSciRef} \\
\midrule
\multirow{2}{*}{0.4B} & $\rho(\text{Ent}, \text{Acc} \mid \text{Surf})$ & $0.150^{*}$ & $0.197^{*}$ & $0.133^{*}$ \\
 & $\rho(\text{Surf}, \text{Acc} \mid \text{Ent})$ & $0.042^{*}$ & $-0.084^{*}$ & $0.080^{*}$ \\
\midrule
\multirow{2}{*}{1.7B} & $\rho(\text{Ent}, \text{Acc} \mid \text{Surf})$ & $0.141^{*}$ & $0.125^{*}$ & $0.151^{*}$ \\
 & $\rho(\text{Surf}, \text{Acc} \mid \text{Ent})$ & $0.049^{*}$ & $0.000$\phantom{$^{*}$} & $0.033$\phantom{$^{*}$} \\

\midrule
\multicolumn{5}{c}{Pythia} \\
\midrule
\multirow{2}{*}{410M} & $\rho(\text{Ent}, \text{Acc} \mid \text{Surf})$ & $0.107^{*}$ & $0.161^{*}$ & $0.091^{*}$ \\
 & $\rho(\text{Surf}, \text{Acc} \mid \text{Ent})$ & $0.069^{*}$ & $-0.050$\phantom{$^{*}$} & $0.070^{*}$ \\
\midrule
\multirow{2}{*}{2.8B} & $\rho(\text{Ent}, \text{Acc} \mid \text{Surf})$ & $0.107^{*}$ & $0.124^{*}$ & $0.114^{*}$ \\
 & $\rho(\text{Surf}, \text{Acc} \mid \text{Ent})$ & $0.076^{*}$ & $-0.008$\phantom{$^{*}$} & $0.051$\phantom{$^{*}$} \\
\midrule
\multirow{2}{*}{6.9B} & $\rho(\text{Ent}, \text{Acc} \mid \text{Surf})$ & $0.126^{*}$ & $0.141^{*}$ & $0.124^{*}$ \\
 & $\rho(\text{Surf}, \text{Acc} \mid \text{Ent})$ & $0.073^{*}$ & $-0.013$\phantom{$^{*}$} & $0.066^{*}$ \\
\midrule
\multirow{2}{*}{12B} & $\rho(\text{Ent}, \text{Acc} \mid \text{Surf})$ & $0.142^{*}$ & $0.148^{*}$ & $0.146^{*}$ \\
 & $\rho(\text{Surf}, \text{Acc} \mid \text{Ent})$ & $0.074^{*}$ & $-0.009$\phantom{$^{*}$} & $0.064^{*}$ \\

\bottomrule
\end{tabular}
\caption{
Results of the partial-correlation analysis.
Here, $\mathrm{Ent}$ and $\mathrm{Surf}$ denote log-transformed entity and surface frequencies, respectively, and $\mathrm{Acc}$ denotes accuracy.
Each value reports a Pearson partial correlation between one log-frequency signal and accuracy while controlling for the other, namely $\rho(\mathrm{Ent}, \mathrm{Acc}\mid \mathrm{Surf})$ and $\rho(\mathrm{Surf}, \mathrm{Acc}\mid \mathrm{Ent})$.
Superscript $^{*}$ indicates that the partial correlation is significantly different from zero ($p<0.01$).
}
\label{tab:partial_correlation}
\end{table}

\subsection{Partial-Correlation Analysis}
Because entity and surface frequencies are correlated, simple correlations cannot determine whether each frequency type has an association with accuracy beyond the other.
We therefore compute Pearson partial correlations between log frequency and accuracy while controlling for the other frequency type.
A partial correlation $\rho(X,Y\mid Z)$ measures the correlation between $X$ and $Y$ after linearly removing the variation explained by a control variable $Z$, equivalently by correlating the residuals of $X$ and $Y$ after regressing both on $Z$.
Specifically, we compute $\rho(\mathrm{Ent}, \mathrm{Acc}\mid \mathrm{Surf})$ to measure the association between entity frequency and accuracy after controlling for surface frequency, and $\rho(\mathrm{Surf}, \mathrm{Acc}\mid \mathrm{Ent})$ for the reverse direction.

\Cref{tab:partial_correlation} summarizes the results for all training-transparent model families.
In the overall and redirect-only subsets, both entity and surface frequencies often retain positive partial correlations with accuracy, indicating that each captures information not fully explained by the other.
In the canonical-only subset, however, the partial correlation for surface frequency is typically close to zero or negative, whereas entity frequency remains consistently positive.
This suggests that, for canonical surface forms, aggregate entity frequency is more informative than the frequency of the canonical surface form alone.
We further verify in Appendix~\ref{appendix:low_frequency_control} that the redirect-only results are not driven solely by extremely low-frequency redirect surface forms.

\subsection{Discussion}
These results are inconsistent with a purely independent surface-specific account.
Accuracy for a target surface form is associated not only with that surface form's own frequency, but also with the aggregate frequency of the corresponding entity.
This pattern is consistent with cross-surface coupling in factual access, rather than independent memorization of each surface form.
The coupling is clearest for canonical surfaces, where surface frequency has little independent association with accuracy once entity frequency is controlled for, whereas entity frequency remains a consistent predictor.

Prior entity-based QA studies commonly evaluate factual memorization through canonical surface forms and relate performance to aggregate entity frequency~\cite{kandpal2023large}.
Our analysis extends this setting by decomposing entity frequency into surface-level frequencies.
The results suggest that this conventional focus on entity frequency remains a useful lens, especially when evaluation uses canonical surface forms, but it also obscures surface-form effects that become visible when alternative names are considered.

As a complementary probe, Appendix~\ref{appendix:entity_linking} reports an entity-linking-style binary QA experiment that directly asks whether a model links two surface forms to the same entity.
Pythia-12B shows only modest balanced accuracy in this probe, suggesting that surface-form equivalence recognition is incomplete and does not by itself explain the category-wise consistency patterns observed in \cref{sec:consistency}.
A fuller account of surface-dependent factual access will require analyses that more directly examine the internal representations and retrieval processes underlying these effects.

\section{Related Work}
\label{sec:related_work}

\paragraph{Memorization in LLMs.}
Analyses of LLM memorization are often divided by whether they focus on exact reproduction or factual generalization~\cite{kandpal2023large}.
One line of work studies \emph{verbatim memorization}, the literal reproduction of training data~\cite{carlini2021extracting,carlini2023quantifying,chen-etal-2024-multi-perspective}, which is closely related to privacy risks and data leakage.
Another line studies \emph{non-verbatim memorization}, where models retain factual associations that can be elicited without reproducing the original training text.
This setting is commonly evaluated through entity-based QA datasets~\cite{kandpal2023large,mallen-etal-2023-trust,maekawa-etal-2024-retrieval}.
Our work belongs to the latter line, focusing on how entity surface forms affect access to memorized factual knowledge.

\paragraph{Entity-based factual memorization.}
\citet{kandpal2023large} extracted entity-based QA pairs from open-domain datasets such as NaturalQuestions~\cite{kwiatkowski-etal-2019-natural} and TriviaQA~\cite{joshi-etal-2017-triviaqa}, showing that facts with low training-data frequency are less likely to be answered correctly.
\citet{elazar2023measuringcausaleffectsdata} used a causal analysis of masked language models to show that simple training-data statistics, such as co-occurrence counts, can affect factual predictions.
\citet{mallen-etal-2023-trust} introduced PopQA and, together with EntityQuestions~\cite{sciavolino-etal-2021-simple}, showed that LLMs struggle with less popular entities, measured by Wikipedia page views.
\citet{maekawa-etal-2024-retrieval} introduced WitQA and further showed that relation frequency also affects factual knowledge memorization.
These studies provide important insights into factors that predict factual QA success, but they typically instantiate each entity with a single canonical surface form.
As a result, they do not separate whether a model has memorized a fact about an entity from whether it can access that fact through a particular entity name.
RedirectQA addresses this gap by pairing the same factual triples with multiple categorized surface forms for each entity.

\paragraph{Robustness and consistency under input variation.}
Robustness and consistency under meaning-preserving input variation have been studied in several settings.
\citet{zheng2024large} investigated robustness to surface-level variations in multiple-choice questions, and \citet{andriushchenko2025does} examined whether safety-aligned LLMs maintain consistent refusal behavior under tense variations.
In QA and factual prediction settings, \citet{ribeiro-etal-2019-red} proposed evaluating models through consistency constraints across related questions, and \citet{elazar-etal-2021-measuring} showed that meaning-preserving paraphrases can still yield inconsistent factual predictions.
These studies primarily examine prompt- or question-level variation.
In contrast, our work isolates variation in the entity mention itself while holding the underlying entity, factual relation, and answer fixed.
This allows us to analyze how factual access differs across naturally occurring categories of entity surface forms, such as aliases, abbreviations, spelling variants, and common errors.

\section{Conclusion}
We introduced \emph{RedirectQA}, an entity-based factual QA dataset that pairs Wikidata factual triples with multiple categorized entity surface forms using Wikipedia redirect information.
Using RedirectQA, we showed that LLM prediction outcomes often change when only the subject entity surface form is changed, indicating that access to memorized factual knowledge is partially surface-dependent.
The inconsistency is category-dependent: models are relatively robust to minor orthographic variations, such as spelling differences, but less consistent for larger lexical variations, such as aliases, alternative names, and abbreviations.
Our frequency analyses further showed that accuracy is associated with both the frequency of a specific surface form and the aggregate frequency of the corresponding entity, suggesting cross-surface coupling in factual access rather than purely independent memorization of each surface form.
Overall, our findings show that evaluating non-verbatim memorization through canonical entity names alone can miss surface-conditioned failures, highlighting the importance of surface-form diversity in factual QA evaluation.

\section*{Limitations}
Our analysis focuses on English factual QA and does not cover multilingual, cross-lingual, or domain-specific settings.
RedirectQA relies on Wikipedia and Wikidata, whose coverage and naming conventions reflect the biases and editorial practices of Wikimedia projects; Wikipedia redirects provide a systematic source of surface forms, but they do not cover all real-world ways of referring to entities.
Our dataset also varies only the subject entity surface form, leaving object-side variation and broader question paraphrasing beyond the main scope.

Although our evaluation uses alias-aware string matching for answer entities, it may still miss semantically correct answers whose surface forms are not included in the acceptable answer set.
Our frequency-based analyses are restricted to training-transparent models and depend on entity-linking quality and the filtered subset of surface instances observed in the relevant corpora.
Finally, our experiments evaluate factual access through QA behavior and frequency correlations, but do not directly probe the internal representations or training dynamics that give rise to surface-dependent access.
Future work could extend RedirectQA to multilingual and domain-specific settings, broaden surface-form resources beyond Wikipedia redirects, and develop more direct analyses of how surface--entity associations are represented and acquired during pretraining.

\section*{Ethical Considerations}
This study uses publicly available data from Wikimedia projects, including Wikipedia, Wikidata, and pageview statistics.
We follow the licenses of the original resources: Wikipedia text is distributed under CC BY-SA 4.0, while Wikidata and pageview statistics are distributed under CC0 1.0.
RedirectQA contains entity names, redirect titles, factual triples, and generated questions derived from these public resources.
It does not include private or newly collected sensitive personal information, although some entities may correspond to public figures already represented in Wikimedia projects.

RedirectQA also uses question templates adapted from PopQA~\cite{mallen-etal-2023-trust}, which is distributed under the MIT license.
To ensure transparency and reproducibility, we make RedirectQA available under the CC BY-SA 4.0 license, following the most restrictive license among the source resources.
Because Wikipedia and Wikidata coverage is not demographically or geographically uniform, RedirectQA may reflect biases present in these resources.
The dataset is intended for evaluating factual QA behavior and should not be used to draw normative conclusions about individuals or groups.

\section*{Acknowledgments}
This work was partially supported by JST SPRING Grant Number JPMJSP2140 and JSPS KAKENHI Grant Number JP23H03458.
Computational resources were provided in part by ``mdx: a platform for building data-empowered society.''

\bibliography{custom}

\appendix

\section{Details on RedirectQA Dataset}
\subsection{Redirect Category Statistics}
\label{sec:appendix_redirect_category_stats}

\Cref{tab:redirect_category_with_count} provides a detailed breakdown of RedirectQA by redirect category.
The Count column reports the number of surface instances assigned to each category.
Because a redirect surface instance may be associated with multiple redirect categories, category-wise counts are not mutually exclusive and should not be summed to recover the total number of redirect surface instances.
Similarly, broad-type counts reported in \cref{sec:dataset_construction} count unique surface instances associated with each type, whereas \cref{tab:redirect_category_with_count} reports counts at the category level.

\begin{table}[t]
\centering
\small
\tabcolsep 3pt
\begin{tabular}{@{}llr@{}}
\toprule
\multicolumn{1}{c}{Type} & \multicolumn{1}{c}{Redirect category} & \multicolumn{1}{c}{Count} \\
\midrule
Canonical & \multicolumn{1}{c}{$-$} & 14,672\\
\midrule
\multirow{20}{*}{Alt./Abbrev.} 
    & \texttt{from birth names} & 1,029 \\
    & \texttt{from short names} & 985 \\
    & \texttt{from alternative names} & 981 \\
    & \texttt{from former names} & 979 \\
    & \texttt{from surnames} & 977 \\
    & \texttt{from abbreviations} & 863 \\
    & \texttt{from initialisms} & 800 \\
    & \texttt{from long names} & 517 \\
    & \texttt{from given names} & 395 \\
    & \texttt{from pseudonyms} & 374 \\
    & \texttt{from personal names} & 371 \\
    & \texttt{from plurals} & 331 \\
    & \texttt{from married names} & 137 \\
    & \texttt{from acronyms} & 122 \\
    & \texttt{from letter–word combinations} & 108 \\
    & \texttt{from technical names} & 87 \\
    & \texttt{to plurals} & 82 \\
    & \texttt{to initialisms} & 74 \\
    & \texttt{from synonyms} & 65 \\
    & \texttt{to acronyms} & 35 \\
\midrule
\multirow{10}{*}{Spell. Var.}  
    & \texttt{from titles without diacritics} & 1,019 \\
    & \texttt{from alternative spellings} & 1,014 \\
    & \texttt{from titles with diacritics} & 998 \\
    & \texttt{from other capitalisations} & 953 \\
    & \texttt{from modifications} & 765 \\
    & \texttt{from ASCII-only titles} & 56 \\
    & \texttt{from stylizations} & 86 \\
    & \texttt{from titles without ligatures} & 61 \\
    & \texttt{to ASCII-only titles} & 35 \\
    & \texttt{from numerals} & 23 \\
\midrule
\multirow{3}{*}{Typ. Err.}       
    & \texttt{from miscapitalisations} & 1,005 \\
    & \texttt{from misspellings} & 978 \\
    & \texttt{from incorrect names} & 974 \\
\bottomrule
\end{tabular}
\caption{
Dataset composition by canonical and redirect surface categories.
The Count column indicates the number of surface instances assigned to each category.
A redirect surface instance may belong to multiple categories, so category counts are not mutually exclusive.
The broad-type counts reported in \cref{sec:dataset_construction} count unique surface instances per type and therefore need not equal the sum of category-level counts.
}
\label{tab:redirect_category_with_count}
\end{table}

\section{Additional Experiments}

\subsection{Preliminary Experiment}
\label{appendix:preliminary_experiment}

As a preliminary diagnostic, we augmented PopQA~\cite{mallen-etal-2023-trust} with Wikipedia redirect information using a procedure similar to that described in \cref{sec:dataset_construction}.
This produced 18,781 surface instances from 4,292 factual triples.
For the consistency analysis, we formed canonical--redirect comparison pairs for which both the canonical and redirect questions were evaluated, yielding 14,489 pairs.

\Cref{tab:redirect_popqa} shows the resulting correctness contingency table for Pythia-12B using the original question template.
Among these canonical--redirect pairs, 23.7\% yielded inconsistent correctness outcomes: the model was correct on one surface form but incorrect on the other.
This preliminary result motivates the more systematic construction of RedirectQA.

\begin{table}[t]
\centering
\small
\begin{tabular}{@{}lrr@{}}
\toprule
& \multicolumn{2}{c}{Canonical surface} \\
\cmidrule(l){2-3}
Redirect surface & Correct & Incorrect \\
\midrule
Correct & 4,285 & 608 \\
Incorrect & 2,821 & 6,775 \\
\bottomrule
\end{tabular}
\caption{
Preliminary consistency analysis on a redirect-augmented version of PopQA using Pythia-12B.
Rows indicate correctness under the redirect surface form, and columns indicate correctness under the canonical surface form.
Counts are canonical--redirect comparison pairs.
In 23.7\% of pairs, the correctness outcome differs across the two surface forms.
}
\label{tab:redirect_popqa}
\end{table}

\subsection{Robustness to Question Templates}
\label{appendix:template_robustness}

In \cref{sec:consistency}, we analyzed prediction consistency using the original template adopted from \citet{mallen-etal-2023-trust}.
To examine whether the observed surface-form effects depend on a particular question wording, we repeat the same analysis using an additional paraphrased template generated by GPT-4o.

\begin{figure*}[t]
\centering
\includegraphics[width=1.0\linewidth]{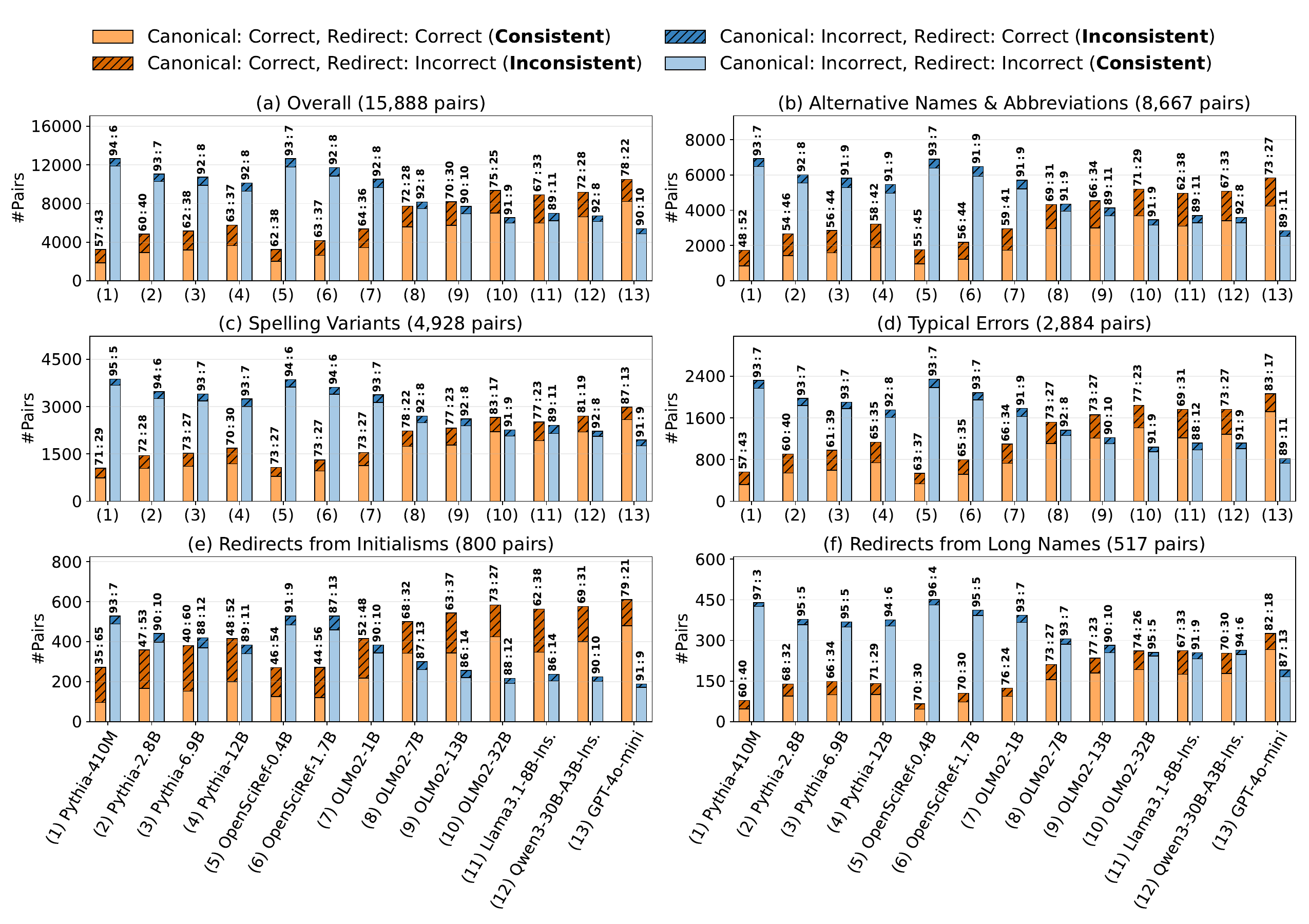}
\caption{
Prediction consistency between canonical and redirect surface forms on RedirectQA using the paraphrased question template.
The plotting convention is the same as in \cref{fig:redirectqa_summary}: light segments indicate consistent correctness outcomes, while dark hatched segments indicate correctness flips.
}
\label{fig:redirectqa_summary_llm_template}
\end{figure*}

\Cref{fig:redirectqa_summary_llm_template} shows the results with the paraphrased template, using the same plotting convention as \cref{fig:redirectqa_summary}.
Although absolute accuracy can vary with question wording, the model-wise consistency patterns and qualitative differences across redirect types largely mirror those obtained with the original template.
This suggests that the main surface-form effects are not artifacts of a single question template.

\subsection{Significance Test and Correlation Results for Transparent Models}
\label{appendix:frequency_correlation}

This section provides supplementary results for the analyses in \cref{sec:analysis}.

\paragraph{Steiger's $Z$-test.}
To test whether the difference between the entity-frequency and surface-frequency correlations reported in \cref{fig:count_vs_score} is statistically meaningful, we conducted Steiger's $Z$-test~\cite{10.1037/0033-2909.87.2.245,hoerger2013zh}, which compares two dependent Pearson correlations that share a common variable.
For the Pythia-12B canonical-only subset, the test confirms that the correlation between entity frequency and accuracy is significantly larger than that between surface frequency and accuracy ($p<0.01$, two-tailed).

\paragraph{Correlation results for other models.}
\Cref{tab:correlation} reports Pearson correlations between log-transformed frequencies and accuracy for all training-transparent models.
Across all models and subsets, both entity and surface frequencies show statistically significant positive correlations with accuracy.
In the canonical-only subset, entity frequency consistently correlates more strongly with accuracy than surface frequency.

\begin{table}[t]
\centering
\small
\tabcolsep 3pt
\begin{tabular}{@{}llrrr@{}}
\toprule
Model & Type & Overall & Canonical & Redirect \\
\midrule

Pythia-410M & Entity & $0.175^{*}$ & $0.212^{*}$ & $0.138^{*}$ \\
 & Surface & $0.155^{*}$ & $0.148^{*}$ & $0.126^{*}$ \\
\midrule
Pythia-2.8B & Entity & $0.179^{*}$ & $0.207^{*}$ & $0.154^{*}$ \\
 & Surface & $0.163^{*}$ & $0.167^{*}$ & $0.116^{*}$ \\
\midrule
Pythia-6.9B & Entity & $0.199^{*}$ & $0.228^{*}$ & $0.173^{*}$ \\
 & Surface & $0.172^{*}$ & $0.182^{*}$ & $0.137^{*}$ \\
\midrule
Pythia-12B & Entity & $0.218^{*}$ & $0.244^{*}$ & $0.196^{*}$ \\
 & Surface & $0.183^{*}$ & $0.197^{*}$ & $0.146^{*}$ \\

\midrule
OpenSciRef-0.4B & Entity & $0.208^{*}$ & $0.228^{*}$ & $0.189^{*}$ \\
 & Surface & $0.151^{*}$ & $0.143^{*}$ & $0.157^{*}$ \\
\midrule
OpenSciRef-1.7B & Entity & $0.202^{*}$ & $0.220^{*}$ & $0.186^{*}$ \\
 & Surface & $0.154^{*}$ & $0.182^{*}$ & $0.115^{*}$ \\

\midrule
OLMo-2-1B & Entity & $0.205^{*}$ & $0.217^{*}$ & $0.196^{*}$ \\
 & Surface & $0.166^{*}$ & $0.138^{*}$ & $0.174^{*}$ \\
\midrule
OLMo-2-7B & Entity & $0.201^{*}$ & $0.203^{*}$ & $0.203^{*}$ \\
 & Surface & $0.195^{*}$ & $0.160^{*}$ & $0.195^{*}$ \\
\midrule
OLMo-2-13B & Entity & $0.188^{*}$ & $0.188^{*}$ & $0.191^{*}$ \\
 & Surface & $0.192^{*}$ & $0.159^{*}$ & $0.193^{*}$ \\
\midrule
OLMo-2-32B & Entity & $0.146^{*}$ & $0.150^{*}$ & $0.145^{*}$ \\
 & Surface & $0.155^{*}$ & $0.104^{*}$ & $0.173^{*}$ \\

\bottomrule
\end{tabular}
\caption{
Pearson correlation coefficients between accuracy and log-transformed entity and surface frequencies.
``Entity'' uses the total frequency aggregated over all observed linked surface forms of an entity, whereas ``Surface'' uses the frequency of the specific surface form.
Superscript $^{*}$ indicates that the correlation is significantly different from zero ($p<0.01$).
}
\label{tab:correlation}
\end{table}

\subsection{Low-Frequency-Controlled Analysis for Redirect Surface Forms}
\label{appendix:low_frequency_control}

\Cref{fig:count_vs_score} shows that redirect surface forms include many extremely low-frequency cases.
This raises the possibility that the redirect-only results in \cref{sec:analysis} are driven primarily by very rare redirect surface forms, rather than by surface-form variation more generally.
To address this concern, we repeat the correlation and partial-correlation analyses on a high-frequency subset of the redirect-only data.
Specifically, we retain only redirect surface instances whose raw surface frequency is greater than 10 in the corresponding corpus.
This filtering is stricter than the main preprocessing, which removes only zero-frequency cases.

\Cref{tab:redirect_highfreq_control} compares the full redirect-only subset with this high-frequency subset.
The full-subset columns reproduce the redirect-only correlations and partial correlations reported in \cref{tab:correlation} and \cref{tab:partial_correlation}, while the high-frequency columns report the same analyses after excluding extremely low-frequency redirect surface instances.
Across all training-transparent models, entity frequency remains significantly associated with accuracy after controlling for surface frequency.
For Pythia and OpenSciRef, the partial correlation of surface frequency becomes smaller and is not statistically significant in the high-frequency subset once entity frequency is controlled for.
For OLMo~2, surface frequency retains a positive partial correlation, but the qualitative pattern remains broadly consistent with the main analysis: removing extremely low-frequency redirect surfaces does not eliminate the entity-frequency signal.
These results suggest that our conclusions are not driven solely by redirect surface forms that appear only a few times in the pretraining corpus.

\begin{table*}[t]
\centering
\small
\begin{tabular}{@{}llllllllll@{}}
\toprule
& & \multicolumn{4}{c}{Full redirect-only subset} 
  & \multicolumn{4}{c}{High-frequency subset} \\
\cmidrule(lr){3-6}
\cmidrule(l){7-10}
Family & Size 
& $\rho_E$ & $\rho_S$ & $\rho_{E\mid S}$ & $\rho_{S\mid E}$ 
& $\rho_E$ & $\rho_S$ & $\rho_{E\mid S}$ & $\rho_{S\mid E}$ \\
\midrule
\multirow{4}{*}{Pythia}
& 410M & $0.138^{*}$ & $0.126^{*}$ & $0.091^{*}$ & $0.070^{*}$ 
       & $0.147^{*}$ & $0.109^{*}$ & $0.108^{*}$ & $0.042$ \\
& 2.8B & $0.154^{*}$ & $0.116^{*}$ & $0.114^{*}$ & $0.051$ 
       & $0.168^{*}$ & $0.108^{*}$ & $0.133^{*}$ & $0.028$ \\
& 6.9B & $0.173^{*}$ & $0.137^{*}$ & $0.124^{*}$ & $0.066^{*}$ 
       & $0.191^{*}$ & $0.133^{*}$ & $0.145^{*}$ & $0.045$ \\
& 12B  & $0.196^{*}$ & $0.146^{*}$ & $0.146^{*}$ & $0.064^{*}$ 
       & $0.218^{*}$ & $0.153^{*}$ & $0.165^{*}$ & $0.053$ \\
\midrule
\multirow{2}{*}{OpenSciRef}
& 0.4B & $0.189^{*}$ & $0.157^{*}$ & $0.133^{*}$ & $0.080^{*}$ 
       & $0.205^{*}$ & $0.136^{*}$ & $0.161^{*}$ & $0.039$ \\
& 1.7B & $0.186^{*}$ & $0.115^{*}$ & $0.151^{*}$ & $0.033$ 
       & $0.205^{*}$ & $0.130^{*}$ & $0.163^{*}$ & $0.033$ \\
\midrule
\multirow{4}{*}{OLMo~2}
& 1B  & $0.196^{*}$ & $0.174^{*}$ & $0.125^{*}$ & $0.084^{*}$ 
      & $0.209^{*}$ & $0.177^{*}$ & $0.140^{*}$ & $0.081^{*}$ \\
& 7B  & $0.203^{*}$ & $0.195^{*}$ & $0.120^{*}$ & $0.106^{*}$ 
      & $0.201^{*}$ & $0.181^{*}$ & $0.127^{*}$ & $0.092^{*}$ \\
& 13B & $0.191^{*}$ & $0.193^{*}$ & $0.106^{*}$ & $0.111^{*}$ 
      & $0.184^{*}$ & $0.180^{*}$ & $0.108^{*}$ & $0.100^{*}$ \\
& 32B & $0.145^{*}$ & $0.173^{*}$ & $0.065^{*}$ & $0.114^{*}$ 
      & $0.142^{*}$ & $0.158^{*}$ & $0.071^{*}$ & $0.100^{*}$ \\
\bottomrule
\end{tabular}
\caption{
Low-frequency-controlled frequency analysis for the redirect-only subset.
The full redirect-only subset uses the same filtering criterion as \cref{sec:analysis}; the high-frequency subset additionally retains only redirect surface instances whose raw surface frequency satisfies $f_{\mathrm{surf}}>10$.
$\rho_E$ and $\rho_S$ denote Pearson correlations between accuracy and log-transformed entity and surface frequencies, respectively.
$\rho_{E\mid S}$ and $\rho_{S\mid E}$ denote Pearson partial correlations, corresponding to $\rho(\mathrm{Ent},\mathrm{Acc}\mid\mathrm{Surf})$ and $\rho(\mathrm{Surf},\mathrm{Acc}\mid\mathrm{Ent})$.
Superscript $^{*}$ indicates that the correlation or partial correlation is significantly different from zero ($p<0.01$).
}
\label{tab:redirect_highfreq_control}
\end{table*}

\subsection{Evaluating Entity Linking Between Surface Forms}
\label{appendix:entity_linking}

As a complementary behavioral probe, we evaluate whether a model can recognize that two surface forms refer to the same entity.
This experiment does not directly reveal internal representations, but provides an additional measure of surface-to-entity linking behavior that may relate to the consistency patterns observed in \cref{sec:consistency}.

We constructed binary questions asking whether two surface forms refer to the same entity.
For positive examples, we used canonical--redirect pairs from RedirectQA and generated category-specific yes/no questions with GPT-4o.
For example, for the redirect category \texttt{from initialisms}, we used the template:
``Is \texttt{<redirect surface>} an initialism for \texttt{<canonical surface>}?''
An example instance is ``Is \textit{NYT} an initialism for \textit{The New York Times}?''

For each positive example, we created two negative examples:
(i) surface-level negatives, obtained by randomly replacing one character in a surface form, and
(ii) semantic negatives, obtained by replacing the entity with a semantically similar but distinct entity retrieved through nearest-neighbor search in the fastText embedding space~\cite{bojanowski-etal-2017-enriching}.
We used the publicly available English model \texttt{cc.en.300.bin},\footnote{\url{https://fasttext.cc/docs/en/crawl-vectors.html}} and measured similarity using squared-L2 distance in the embedding space.
Thus, the evaluation set has a 1:2 ratio of positive to negative examples.

\Cref{tab:el_qa_accuracy} reports raw and balanced accuracy for Pythia-12B across redirect types and selected categories.
Balanced accuracy is computed as the average of positive-example accuracy and negative-example accuracy, treating the two negative types as a single negative class.
Because the evaluation set has a 1:2 positive-to-negative ratio, raw accuracy can be affected by the larger number of negative examples.
Balanced accuracy therefore provides a more appropriate summary under this label imbalance.

Overall balanced accuracy is 0.522, indicating only a modest ability to recognize that two surface forms refer to the same entity.
The model is substantially more accurate on negative examples than on positive examples, suggesting that it is better at rejecting mismatched surface forms than at affirming true surface-form equivalences.
Across broad redirect types, balanced accuracy is highest for \emph{Alternative Names and Abbreviations} and close to the balanced-accuracy random baseline of 0.5 for \emph{Spelling Variants} and \emph{Typical Errors}.
For the two selected subcategories analyzed in \cref{sec:consistency}, \texttt{from initialisms} and \texttt{from long names} show similar balanced accuracies, despite exhibiting different factual QA consistency patterns in \cref{fig:redirectqa_summary}.
This suggests that the binary surface-linking probe alone cannot explain the category-wise differences in factual QA consistency.
A deeper analysis of the mechanisms underlying surface-dependent factual access remains an important direction for future work.

\begin{table}[t]
\centering
\small
\tabcolsep 3pt
\begin{tabular}{@{}lrrrr@{}}
\toprule
Redirect Category/Type & Raw & Bal. & Pos. & Neg. \\
\midrule
Overall & 0.585 & 0.522 & 0.334 & 0.711 \\
\midrule
Alt./Abbrev. & 0.587 & 0.537 & 0.386 & 0.688 \\
Spell. Var. & 0.574 & 0.507 & 0.305 & 0.708 \\
Typ. Err. & 0.604 & 0.506 & 0.212 & 0.801 \\
\midrule
\texttt{from initialisms} & 0.587 & 0.564 & 0.495 & 0.633 \\
\texttt{from long names} & 0.592 & 0.559 & 0.461 & 0.657 \\
\bottomrule
\end{tabular}
\caption{
Results of the entity-linking-style binary QA task with Pythia-12B.
For each positive canonical--redirect pair, we include two negative examples: one surface-level negative and one semantic negative.
Raw accuracy is computed over all examples.
Balanced accuracy averages positive accuracy and negative accuracy, where the two negative types are treated as a single negative class.
\textsc{Pos.} and \textsc{Neg.} denote accuracy on positive and negative examples, respectively.
}
\label{tab:el_qa_accuracy}
\end{table}

\section{Data, Models, and Software}

\subsection{Data}
\begin{description}
\item[Wikimedia Dumps] provided by the Wikimedia Foundation. License: CC BY-SA 4.0 (Wikipedia text), CC0 1.0 (Wikidata and pageviews).
\url{https://dumps.wikimedia.org/}.

\item[PopQA] created by \newcite{mallen-etal-2023-trust}. License: MIT.
\url{https://github.com/AlexTMallen/adaptive-retrieval}

\item[The Pile] created by \newcite{gao2020pile800gbdatasetdiverse}.
The Pile is a composite dataset consisting of multiple component datasets; licensing and usage terms vary by component.
\url{https://pile.eleuther.ai/}.

\item[OLMo Mix 1124] created by \newcite{olmo20242olmo2furious}.
License: ODC-By v1.0; use is also subject to Common Crawl's Terms of Use.
\url{https://huggingface.co/datasets/allenai/olmo-mix-1124}.
\end{description}

\subsection{Models}
\begin{description}
\item[Pythia] created by \newcite{biderman2023pythia}.
License: Apache-2.0.
\url{https://huggingface.co/collections/EleutherAI/pythia-scaling-suite}. 

\item[OLMo 2] created by \newcite{olmo20242olmo2furious}.
License: Apache-2.0.
\url{https://huggingface.co/collections/allenai/olmo-2}

\item[open-sci-ref-0.01 Pile] created by \newcite{nezhurina2025opensciref001openreproduciblereference}. 
License: Apache-2.0.
\url{https://huggingface.co/collections/open-sci/open-sci-ref-001-pile}

\item[Llama 3.1] created by \newcite{grattafiori2024llama3herdmodels}.
License: Meta Llama 3.1 Community License.
\url{https://www.llama.com/llama3_1/}

\item[Qwen3] created by \newcite{yang2025qwen3technicalreport}.
License: Apache 2.0.
\url{https://huggingface.co/collections/Qwen/qwen3}

\item[GPT-4o] created by \newcite{openai2024gpt4o}.
License: Proprietary; access governed by OpenAI's Terms of Use.

\item[GPT-4o-mini] created by \newcite{openai2024gpt4omini}.
License: Proprietary; access governed by OpenAI's Terms of Use.

\item[fastText English word vectors (cc.en.300.bin)] created by \newcite{bojanowski-etal-2017-enriching}.
License: CC BY-SA 3.0.
\url{https://fasttext.cc/docs/en/crawl-vectors.html}. 

\end{description}

\subsection{Software}
\begin{description}
\item[DBpedia Spotlight] created by \newcite{mendes2011dbpedia}. 
License: Apache-2.0.
\url{https://github.com/dbpedia-spotlight/dbpedia-spotlight}. 

\item[fastText] created by \newcite{bojanowski-etal-2017-enriching}. 
License: MIT.
\url{https://github.com/facebookresearch/fastText}. 

\end{description}

\end{document}